\documentclass[11pt]{article}
\usepackage[preprint]{acl}
\usepackage{amsmath}
\usepackage{times}
\usepackage{latexsym}
\usepackage{graphicx}
\usepackage{lipsum} 
\usepackage{hyperref}
\usepackage[table,xcdraw]{xcolor}

\usepackage[T1]{fontenc}

\usepackage{algorithm}
\usepackage[utf8]{inputenc}
\usepackage{algpseudocode}
\usepackage{microtype}
\usepackage{geometry}

\usepackage{inconsolata}
\usepackage{enumitem}

\usepackage{graphicx}

\usepackage[utf8]{inputenc} 
\usepackage[T1]{fontenc}    
\usepackage{url}      
\usepackage{tabularx}
\usepackage{booktabs}       
\usepackage{amsfonts}       
\usepackage{nicefrac}       
\usepackage{microtype}      
\usepackage{multirow}
\usepackage{graphicx}
\usepackage{pifont}
\usepackage{makecell}
\usepackage{enumitem}
\usepackage{array}
\usepackage{colortbl}
\usepackage{amssymb}  
\usepackage{bbding}    
\usepackage{wasysym}   
\usepackage{etoolbox}
\usepackage{listings}
\usepackage[most]{tcolorbox}
\tcbuselibrary{skins, breakable}
\usepackage{fancyvrb}
\definecolor{promptgray}{RGB}{245, 245, 245}

\definecolor{darkgreen}{RGB}{0,100,0}
\definecolor{mia-green}{RGB}{78, 205, 196}
\definecolor{qwen-red}{RGB}{255, 107, 107}
\definecolor{mia-rag-green}{RGB}{39, 174, 96}
\definecolor{baseline-gray}{RGB}{149, 165, 166}
\definecolor{darkblue}{rgb}{0, 0, 0.5}
\definecolor{poscolor}{RGB}{170,70,70}
\definecolor{negcolor}{RGB}{70,130,100}
\hypersetup{colorlinks=true, citecolor=darkblue, linkcolor=darkblue, urlcolor=darkblue}
\setlist[itemize]{nosep,leftmargin=*}
\setlist[enumerate]{nosep,leftmargin=*}
\definecolor{arrowRed}{RGB}{220, 38, 38}
\definecolor{bgRed}{RGB}{254, 226, 226}

\definecolor{arrowGreen}{RGB}{5, 150, 105}
\definecolor{bgGreen}{RGB}{209, 250, 229}

\newtcolorbox{promptbox}{
  enhanced,
  breakable,
  colback=black!2,
  colframe=black!35,
  boxrule=0.6pt,
  arc=2mm,
  boxsep=4pt,
  left=9pt,right=9pt,top=8pt,bottom=8pt,
  borderline west={2.0pt}{0pt}{black!45},
  fontupper=\footnotesize\normalfont, 
  before skip=0.6\baselineskip,
  after skip=0.6\baselineskip,
  before upper={
    \setlength{\parindent}{0pt}
    \setlength{\parskip}{0.25em} 
  },
}

\newcommand{\PromptLabel}[1]{\textbf{\textsc{#1}}}

\newcommand{\Var}[1]{\texttt{\{#1\}}}

\DefineVerbatimEnvironment{PromptJSON}{Verbatim}{
  fontsize=\footnotesize,
  baselinestretch=0.95
}

\newcounter{rqsection}

\newcounter{rqsous}[rqsection]

\title{Query-focused and Memory-aware Reranker for Long Context Processing}
\author{
Yuqing Li$^{1,2}$\thanks{Equal contribution.} \;
Jiangnan Li$^{3}$\footnotemark[1] \;
Mo Yu$^{3}$\footnotemark[1] \;
Yanyu Chen$^{4}$ \;
Guoxuan Ding$^{1,2}$ \; \\
\textbf{Zheng Lin}$^{1,2}$\thanks{Corresponding author.} \; 
\textbf{Wei Zhang}$^{4}$ \;
\textbf{Jie Zhou}$^{3}$  \\
$^{1}$Institute of Information Engineering, Chinese Academy of Sciences \\
$^{2}$School of Cyber Security, University of Chinese Academy of Sciences \\
$^{3}$Pattern Recognition Center, WeChat AI, Tencent
$^{4}$East China Normal University\\
\texttt{liyuqing@iie.ac.cn}\quad \texttt{\{jiangnanli,moyumyu\}@tencent.com}
}

\usepackage{textcomp}
\DeclareUnicodeCharacter{266B}{\textmusicalnote}

\begin{document}
    \maketitle
\begin{abstract}
Built upon the existing analysis of retrieval heads in large language models, we propose an alternative reranking framework that trains models to estimate passage–query relevance using the attention scores of selected heads.
This approach provides a listwise solution that leverages the holistic information within the entire candidate shortlist during ranking. At the same time, it naturally produces continuous relevance scores, enabling training on arbitrary retrieval datasets without requiring Likert-scale supervision.
Our framework is lightweight and effective, requiring only small-scale models (\emph{e.g.}, 3B parameters) to achieve strong performance.
Extensive experiments demonstrate that our method outperforms existing state-of-the-art pointwise and listwise rerankers across multiple domains, including Wikipedia and long narrative datasets. It further establishes a new state-of-the-art on the LoCoMo benchmark that assesses the capabilities of dialogue understanding and memory usage.
We further demonstrate that our framework supports flexible extensions. For example, augmenting candidate passages with contextual information further improves ranking accuracy, while training attention heads from middle layers enhances efficiency without sacrificing performance.  
\footnote{The models are available at \url{https://huggingface.co/MindscapeRAG/QRRanker}}

\end{abstract}

\section{Introduction}

Embedding Models, especially those built on top of LLMs, achieved successes and enabled generators (RAG) and agents to work with long inputs or large input corpora efficiently~\cite{zhang2025qwen3,zhao2025kalm,babakhin2025llamaembednemotron8buniversaltextembedding,li2025mindscape}.
However, embeddings also have limitations, as theoretically proved and empirically illustrated by \cite{weller2025theoretical}. They reveal a "geometric bottleneck" where fixed-dimensional vectors fail to encode the combinatorial complexity of query-document interactions. 
Furthermore, the inductive bias of the similarity measure limits the applicable domains where other types of relationships are required to recall, \emph{e.g.}, causality, associations, and analogy.

A long line research applies an additional reranker module on the shortlist returned from embedding models to resolve this challenge.
The rerankers use larger models, more powerful representations (like cross-attention).
The fast development of LLMs boosts many LLM-based reranker releases to benefit from the reasoning capabilities of LLMs~\cite{zhang2025qwen3,sun2025grouprank,liu2025reasonrank,pradeep2023rankzephyr}.
These rerankers can adopt either pointwise or listwise formulations.
Pointwise lost the global view of the shortlist, but can give scores.
Listwise approaches, on the other hand, directly inherit the long-context reasoning and text generation ability of the backbone LLMs, which takes a holistic view of the shortlist, but the next-token prediction limits the prediction of fine-grained scores, and the predicted float numbers cannot always accurately reflect the true confidence~\cite{liu2025uncertainty,lin2024just}.
As a result, they adopt a Likert rating regime, asking the models to output a five-point or ten-point scale score for each input document, which limits the available training data.

In this work, we propose an alternative solution built upon the existing analysis of retrieval heads in LLMs~\cite{Wu2024RetrievalHM,zhang2025query}.
These works identify two related types of heads: retrieval heads and Query-focused Retrieval (QR) heads. Both refer to attention heads whose attention patterns reflect retrieval behaviors.
Specifically, when concatenating long contexts of relevant and distractor passages with the query, these heads are defined as those that put significant attention weights on the relevant passages, so as the ranks of attention weights correlate with the ranks of relevance.

While existing works mainly focus on probing and understanding the functions of such heads, our work moves one step further by training LLMs to optimize the ranking accuracy of a small set of retrieval heads. In this way, we achieve an LLM-ranker that is optimized to rank passages with attention weights. This resulted listwise solution, named \textbf{QRRanker}, can naturally work with continuous relevance scores without the limitation of Likert-scale supervision, hence can be trained on arbitrary retrieval datasets.

Our QRRanker enjoys several good properties in practice. First, the retrieval heads can be effectively trained even when the backbone has a relatively small scale, \emph{e.g.}, 3B parameters. This allows the listwise approach to run with improved efficiency.
Second, it is easy to enhance the input candidate passages with their global context with efficiency, by prepending the shared contextual information to the ground of candidates during training, which is essential for long narrative understanding.
Finally, we observed that our QRRanker is quite robust to the selection of heads, and training with heads from layers in the middle would result in no performance drop. This allows us to take off the higher layers of the LLMs during training and inference, which can greatly reduce the latency of the model.


Experiments across diverse retrieval settings, including long narrative QA (NarrativeQA, DetectiveQA),
long-term dialogue memory (LoCoMo), wikipedia multi-hop QA (MuSiQue, HotpotQA), and reasoning-intensive retrieval (BRIGHT), demonstrate the effectiveness of QRRanker. As a versatile reranking framework, QRRanker outperforms strong general-purpose pointwise and listwise rerankers, such as Qwen-Reranker and ReasonRank, and also improves over domain-specific retrieval or memory methods, including HippoRAG-v2~\citep{gutierrez2025} for Wikipedia QA and recent memory-enhanced approaches~\citep{li2025memos,rasmussen2025zep} on LoCoMo. Additional comparisons with  retrieval agents~\cite{li2026beyond,nvidia2026nemoretriever} further suggest that QRRanker can serve as an efficient one-step alternative to iterative retrieval pipelines.
\section{Related Work}

\paragraph{Reranking} 
Reranking methods are commonly built on bi-encoder or cross-encoder architectures~\citep{koch2015siamese,cross-encoder}. Bi-encoders retrieve candidates using reusable document embeddings~\citep{zhang2025qwen3}, but suffer from the ``geometric bottleneck'' and cannot fully model fine-grained query-document interactions. Cross-encoders address this by jointly encoding each query-document pair with cross-attention, but their high cost limits them to reranking the top-$n$ candidates retrieved by bi-encoders.
LLM-based rerankers are typically categorized into pointwise and listwise methods. Pointwise methods score documents independently and are widely used in practice~\citep{prp,rankgpt,liu2025reasonrank,rankr1}, \emph{e.g.}, Qwen3~\citep{zhang2025qwen3}, Jina, mGTE~\citep{mGTE}, and BGE-m3~\citep{m3}; however, independent scoring limits their ability to capture global information across the candidate list. Listwise methods concatenate multiple documents and generate rankings directly~\citep{pradeep2023rankvicuna,pradeep2023rankzephyr}, with recent variants using reasoning or RL for stronger performance~\citep{rankgpt,liu2025reasonrank,tongsearch,sun2025grouprank}. However, they often require ranking-specific supervision and suffer from unstable generation formats, especially when reasoning traces are introduced.

Recent studies show that LLMs inherently possess retrieval ability, and retrieval-related attention heads can be extracted or modulated for retrieval~\citep{Wu2024RetrievalHM,zhang2025query,lee2025seal}. SEAL~\citep{lee2025seal} learns lightweight scaling factors for attention heads or channels to enhance long-context retrieval. In contrast, QRRanker trains selected query-focused retrieval heads as an explicit listwise reranking scorer, producing fine-grained document scores with query-positive-negative supervision.
\paragraph{Memory Utilization}
Memory construction and utilization have been widely studied to alleviate long-context processing challenges. Existing work builds global memories for long-story understanding~\citep{tom25zhou,kovcisky2018narrativeqa,xu2025detectiveqa,wu2025sitemb,li2025mindscape}, and designs graphs~\citep{jiang2026synapse,xu2025mem,rasmussen2025zep,hu2026memory,hu2026evermemos,hgmem26zhou}, trees~\citep{li2026timem}, or memory systems~\citep{chhikara2025mem0,li2025memos,nan2025nemori,tao2026membox,zou2026mem} to retrieve relevant dialogue histories, events, personas, and chunks. In contrast to increasingly complex memory management, we show that stronger retrieval over simply constructed memories can be an effective alternative.
\begin{figure}
    \centering
    \includegraphics[width=0.99\linewidth]{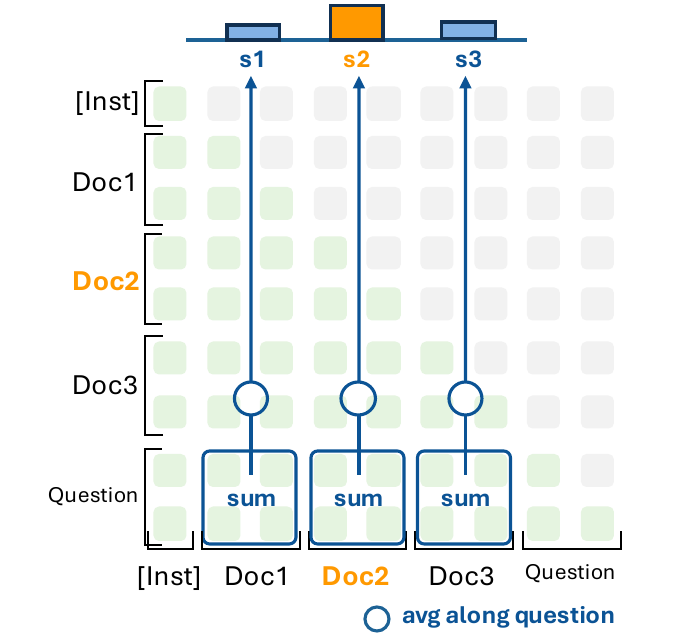}
    \caption{The retrieval score and QR score are computed based on the attention score of a (QR) attention head. In this figure, Doc2 is the gold document (chunk).}
    \label{fig:qr_score}
\end{figure}

\section{Preliminaries: QR-head}\label{preliminaries}

\begin{figure*}
    \centering
    \includegraphics[width=0.95\linewidth]{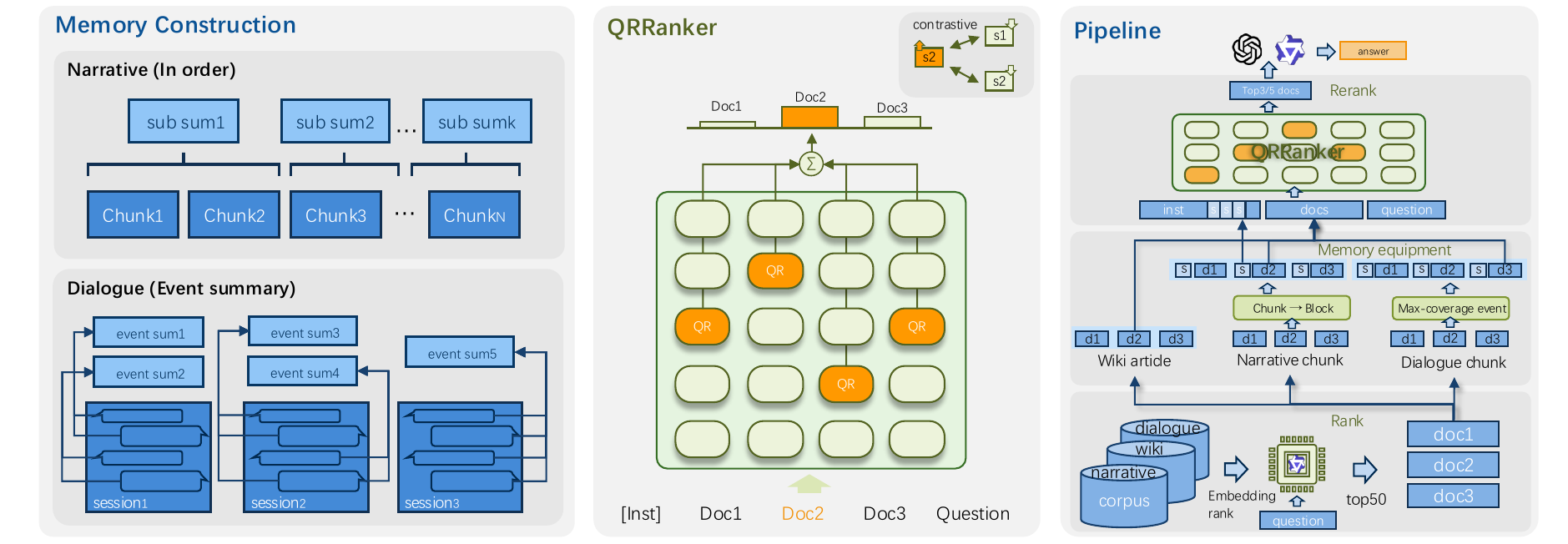}
    \caption{The structure of QRRanker is illustrated in the middle, where the highlighted heads are QR heads for document scoring. As QRRanker can be aware of memory enhancement to capture more contextual information, we can construct memories for narratives and dialogues, which is shown on the left. The right part demonstrates the rank-rerank pipeline of qa for narratives/wiki/dialogues, which involves no sophisticated design.}
    \label{fig:main}
\end{figure*}
We first introduce Query-Focused Retrieval heads (QR-heads). Prior studies show that some self-attention heads act as retrievers by attending to context spans relevant to the question during question encoding~\citep{Wu2024RetrievalHM,zhang2025query}. \citet{zhang2025query} define such heads as QR-heads and identify them using the QR score.

Formally, given a question $Q$ and its context $C=[c_0,c_1,\ldots,c_n]$, let $G=[c_{g0},\ldots,c_{gm}]$ denote the gold chunks. For an attention head $h$, we denote its attention from $Q$ to chunk $c_i$ as $A^{Q\rightarrow c_i}_{h}\in \mathbb{R}^{|Q|\times |c_i|}$. The QR score of $h$ is computed by aggregating its attention to gold chunks:
\begin{equation}
    \texttt{QRScore}_{h}=\frac{1}{|Q|}\sum_{c_i\in G}\sum_{w_q\in Q}\sum_{w_c\in c_i} A^{Q\rightarrow c_i}_{h}[w_q,w_c],
    \label{qr_computing}
\end{equation}
where $w_q$ and $w_c$ denote tokens in $Q$ and $c_i$, respectively. A higher QR score indicates stronger attention to gold chunks. Following \citet{zhang2025query}, we compute the average QR score of each head on a seed set, rank all heads, and select the top 16 as QR-heads $H_{QR}$. In this work, we select QR-heads for \texttt{Qwen3-4B-Instruct-2507}~\citep{yang2025qwen3} using 1,000 random samples from NarrativeQA training split. We also apply the same selection procedure to \texttt{Llama-3.2-3B-Instruct}~\cite{grattafiori2024llama} on 1,000 random samples from MuSiQue in Appx.~\ref{appx:llama_results}, showing that the selected heads are effective beyond a single model or data.

Given the selected QR-heads, the score of a chunk $c_i$ is computed by aggregating its question-to-chunk attention over $H_{QR}$, analogous to Eq.~\ref{qr_computing}.

\section{Method}

QRRanker is a listwise reranking method that processes all candidate documents in a single inference pass.
Unlike generation-based listwise rerankers, it only prefills the question-document prompt and derives document scores from attention patterns, avoiding generation-format errors and reducing inference cost. Since precomputed QR-heads may vary across tasks~\citep{zhang2025query}, we further train them with a dedicated pipeline: constructing listwise training instances and optimizing the selected heads with a contrastive ranking objective.

\subsection{Data Construction for QR Training}

\subsubsection{Listwise Training Instances}
We build a unified training set by combining MuSiQue~\cite{trivedi2022musique} and NarrativeQA~\cite{kovcisky2018narrativeqa}. We first determine evidence chunks for each question. For MuSiQue, we directly use the official supporting facts in the original annotations as evidence. For NarrativeQA, since gold chunks are not provided, we follow~\citet{li2025mindscape} to construct \emph{silver} evidence chunks.

After establishing the evidence, we retrieve a top-$50$ candidate set for each question using Qwen3-Embedding-8B and form a listwise instance by labeling retrieved candidates that match the pre-constructed evidence as positive, while treating the remaining retrieved candidates as negatives.


Optionally, we build a \emph{summary prefix} by mapping the retrieved chunks to their corresponding summaries and prepending these summaries to the chunk list, \emph{i.e.}, $X=[M;C]$. The detailed construction procedure for NarrativeQA is provided in alg.~1 in Appx.~\ref{appx:build_train_instances}. MuSiQue follows the same pipeline, except that relevant evidence is taken directly from its official supporting facts. We describe the summary construction process in the next subsection.

\subsubsection{Summary Construction}
To provide high-level semantic guidance and support long-context narrative understanding, we construct summaries as auxiliary memory context. When used, summaries are prepended as a global prefix to the retrieved chunk list, so the model can leverage both coarse-grained context and fine-grained evidence.
We explore two complementary strategies for constructing summaries.

\paragraph{Block-based Summary.}
For long narrative books, we employ a block-based summarization strategy that respects the temporal flow of the narrative. Specifically, we segment each book into blocks (20 consecutive chunks per block) and generate a corresponding summary for each. This process is detailed in Appx.~\ref{appx:prompt_block_sum}.

\paragraph{Event-centric Summary.}
For dialogue-based data, we extract structured events from conversations and form an event-centric summary. Each event is represented by a short description and is linked to its source utterances, enabling traceability to the original dialogue. (Details in    Appx.~\ref{appx:prompt_event_sum}).

\subsection{QR Training}

Obtaining QR heads precomputed by the QR score mentioned in \textsection~\ref{preliminaries}, our training scheme focuses on training these heads. For a question $Q$ and the top 50 candidate documents $C=[c_1,...,c_{50}]$ ranked by a retriever (\emph{e.g.}, embedding models like Qwen3-Embedding), where gold (positive) documents are $G=[c_{g0},..,c_{gm}]$, the prompt input to QRRanker is constructed by concatenating $C$ and $Q$ in order with some instructions: $\texttt{P}=\text{Inst}(C,Q)$, where the instruction template is provided in Appx.~\ref{appx:prompt_inst}.

The prompt \texttt{P} is fed into the model, and in every attention head, the attention score is computed as $A^{\texttt{P}\rightarrow\texttt{P}}_{h}$. We locate the position of $Q$ and $c_{i}\in C$ and take out the query-focused part $A^{Q\rightarrow c_{i}}_{h}$. The retrieval score of the passage $c_{i}$ computed by the QR head $h\in H_{QR}$ is:
\begin{equation}
    s_{c_{i}}^{h}=\frac{1}{|Q|}\sum_{i\in c_i}\sum_{j\in Q}A^{Q\rightarrow c_{i}}_{h}[i,j],
\end{equation}
where the score computing is illustrated in Fig.~\ref{fig:qr_score}. Then, the final retrieval score is obtained by summing up all scores provided by QR heads: $s_{c_{i}}=\sum_{h\in H_{QR}}{s_{c_{i}}^{h}}$. Additionally, $s_{c_{i}}^{h}$ can also be computed by aggregating the maximum attention item, like used in approaches like ColBERT~\cite{khattab2020colbert}, which achieves similar performance, so we do not discuss it here. 

We then optimize the document scores $S=[s_{c_1},...,s_{c_{50}}]$ utilizing the sample-level contrastive loss. In a conventional contrastive scene, the score $s_{c_{i}}$ is stably ranged in [0, 1], while, in our case, $s_{c_{i}}$ can be affected by tokens in the instruction (\emph{e.g.}, the head's sensitivity to attention sink), which may lead to an unstable range for samples. Therefore, the temperature may not be suitable for scaling the score. To this end, we normalize the score with the max-min norm, which can be formed as:
\begin{equation}
    S = \frac{scale\times(S-\text{min}(S))}{\text{max}(S)-\text{min}(S)},
\end{equation}
where $scale$ is a factor to scale the range to [0, $scale$] for stability. 

The original contrastive loss samples one positive document at a time; however, the top 50 documents may contain more than one positive document. It can be suboptimal if we follow the original setting, as unselected positive documents are ignored. We adopt a group version of contrastive loss to simultaneously optimize them:
\begin{equation} \mathcal{L}=-\frac{1}{|G|}\sum_{c_{p}\in G}\text{log}\frac{\tau(s_{c_{p}})}{\tau(s_{c_{p}})+\sum_{c_{n}\in C\setminus G}\tau(s_{c_{n}})}, \end{equation} where $\tau$ denotes the exponential function.
The objective above treats each positive document as an independent sub-sample and averages the loss inside the sample. For the dataset, the objective aligns with conventional contrastive loss.

As our QRRanker can be made memory-aware to incorporate broader contextual information, during QR training, we optionally prepend a memory prefix $M$ (\emph{e.g.}, summaries mapped from the retrieved chunks) before the candidate list $C$. The resulting prompt to QRRanker is constructed as $\texttt{P}=\text{Inst}(M,C,Q)$.

\newcommand{\blue}[1]{\textcolor[HTML]{6E96B4}{#1}}
\begin{table*}[t]
\centering
\small
\setlength{\tabcolsep}{3.5pt}
\renewcommand{\arraystretch}{1.15}
\resizebox{\textwidth}{!}{
\begin{tabular}{l|ccc|ccc|ccc|ccc|ccc}
\toprule
\multirow{4}{*}{\textbf{Methods}}
& \multicolumn{6}{c|}{\textbf{Wikipedia QA}}
& \multicolumn{6}{c|}{\textbf{Story QA}}
& \multicolumn{3}{c}{\textbf{Overall}} \\
& \multicolumn{3}{c|}{\textbf{Musique}}
& \multicolumn{3}{c|}{\textbf{HotpotQA}}
& \multicolumn{3}{c|}{\textbf{NarrativeQA}}
& \multicolumn{3}{c|}{\textbf{DetectiveQA}}
& \multicolumn{3}{c}{\textbf{Avg@k}} \\
\cmidrule(lr){2-4}\cmidrule(lr){5-7}\cmidrule(lr){8-10}\cmidrule(lr){11-13}\cmidrule(lr){14-16}
& \textbf{R@3} & \textbf{R@5} & \textbf{R@10}
& \textbf{R@3} & \textbf{R@5} & \textbf{R@10}
& \textbf{R@3} & \textbf{R@5} & \textbf{R@10}
& \textbf{R@3} & \textbf{R@5} & \textbf{R@10}
& \textbf{avg@3} & \textbf{avg@5} & \textbf{avg@10} \\
\midrule

\multicolumn{16}{c}{\emph{\underline{Embedding Methods}}}\\
Qwen3-Embedding-4B & 51.56 & 59.83 & 69.88 & 78.84 & 86.16 & 92.33 & 12.57 & 18.33 & 28.08 & 19.25 & 26.17 & 37.04 & 40.56 & 47.62 & 56.83 \\
Qwen3-Embedding-8B & 54.35 & 62.55 & 72.47 & 82.85 & 89.05 & 95.15 & 14.98 & 20.92 & 32.39 & 12.84 & 20.00 & 31.17 & 41.25 & 48.13 & 57.80 \\
SFT-Embedding-8B   & 45.11 & 52.93 & 62.03 & 82.36 & 88.63 & 94.19 & 21.31 & 29.77 & 44.17 & 19.84 & 27.59 & 39.00 & 42.16 & 49.73 & 59.85 \\
\midrule

\multicolumn{16}{c}{\emph{\underline{Reranking Methods}}}\\
HippoRAG-v1  & -- & 53.20 & -- & -- & 90.40 & -- & -- & -- & -- & -- & -- & -- & -- & -- & -- \\
HippoRAG-v2  & -- & \underline{74.70} & -- & -- & \underline{96.30} & -- & -- & -- & -- & -- & -- & -- & -- & -- & -- \\
Qwen-Reranker-4B (out-of-box) & 57.60 & 66.37 & 74.26 & 89.80 & 94.15 & 96.75 & 20.83 & 28.25 & 41.98 & 23.42 & 30.50 & 42.09 & 47.91 & 54.82 & 63.77 \\
Qwen-Reranker-4B (trained)    & 61.60 & 69.71 & 77.49 & 89.35 & 93.95 & \underline{96.90} & 25.84 & 35.05 & 49.62 & 29.67 & 38.92 & 51.25 & 51.61 & 59.41 & \underline{68.82} \\
GroupRank-32B$^{*}$           & 55.49 & 65.08 & 73.07 & 82.45 & 90.60 & 94.50 & 23.98 & 33.76 & 48.83 & 29.34 & 39.21 & 51.38 & 47.82 & 57.16 & 66.95 \\
ReasonRank-32B          & 57.33 & 64.58 & 72.26 & \underline{92.90} & 95.65 & 96.85 & \underline{28.58} & \underline{37.11} & \underline{50.71} & \textbf{36.00} & \textbf{42.84} & \underline{51.83} & \underline{53.70} & \underline{60.05} & 67.91 \\
QRHeads-4B (out-of-box)       & \underline{63.12} & 71.22 & \underline{78.99} & 90.20 & 94.80 & \underline{96.90} & 24.28 & 33.44 & 48.89 & 23.71 & 32.89 & 45.58 & 50.33 & 58.09 & 67.59 \\
\midrule

\textbf{Our QRRanker-4B} & \textbf{70.19} & \textbf{77.37} & \textbf{82.13}
& \textbf{95.05} & \textbf{96.90} & \textbf{97.70}
& \textbf{29.11} & \textbf{38.89} & \textbf{54.93}
& \underline{32.22} & \underline{41.32} & \textbf{53.76}
& \textbf{56.64} & \textbf{63.62} & \textbf{72.13} \\
\bottomrule
\end{tabular}
}
\caption{Retrieval and Rerank performance measured by Recall@\{k\}. The first-stage retriever R@50 ceiling is 86.17\% (Musique), 97.85\% (HotpotQA), 79.90\% (NarrativeQA), and 63.85\% (DetectiveQA). `--' indicates the metric is not reported in the corresponding paper. For Wikipedia QA, we rerank the top-50 candidates retrieved by Qwen3-Embedding-8B; for Story QA, we rerank the top-50 candidates retrieved by SFT-Embedding-8B. DetectiveQA scores are averaged over English and Chinese sets. Bold and underline denote the best and second-best results, respectively. $^{*}$ For fairness, all rerankers are evaluated with a single run.}
\label{tab:longctx_recall_v2}
\end{table*}

\begin{table}[ht]
\centering
\scriptsize
\renewcommand{\arraystretch}{1.0}
\setlength{\tabcolsep}{7.5pt}
\begin{tabular}{l ccc}
\toprule
\textbf{Methods}& \textbf{R@3} & \textbf{R@5}  & \textbf{R@10} \\
\midrule
Qwen3-Emb-8b  & 58.61 & 67.67 &79.15\\
\rowcolor{gray!20}SFT-Emb-8b   & 76.01 & 83.10 &90.15 \\
\midrule
Qwen-Reranker-4B (out-of-box) & 76.15 & 83.02 & 90.15  \\
Qwen-Reranker-4B (trained) &79.17 & 85.51 & 90.74 \\
GroupRank-32B & 77.99 & 82.94 &88.14 \\
ReasonRank-32B & 82.49 & 86.83 & 92.45 \\
QRHeads (out-of-box) & 85.93 & 90.35 &94.86 \\
\rowcolor{gray!20}\textbf{QRRanker} (ours) & \textbf{87.34} & \textbf{91.32}  & \textbf{95.01}\\
\rowcolor{gray!20}\textit{Improvement vs. SFT-Emb} & \textcolor{red}{+11.33} & \textcolor{red}{+8.22} &  \textcolor{red}{+4.86} \\
\bottomrule
\end{tabular}
\caption{Retrieval and Rerank performance on LoCoMo.  The first-stage retriever R@50 ceiling is 97.87\%}
\label{tab:locomo_retrieval_results}
\end{table}

\section{Experimental Setup}
\subsection{Datasets}

To evaluate QRRanker across diverse retrieval settings, we conduct experiments on benchmarks spanning Wikipedia multi-hop QA, long-context story QA, and dialogue memory.
\paragraph{Wikipedia Multi-hop QA} 
For fact-based multi-hop retrieval, we evaluate on \textbf{HotpotQA}~\cite{yang2018hotpotqa} and \textbf{MuSiQue}~\cite{trivedi2022musique}. To ensure a fair comparison, we adopt the corpus and test splits provided by \textbf{HippoRAG}~\cite{gutierrez2025}, maintaining consistency in the candidate passage pool.
\paragraph{Long-context Story QA} 
We utilize datasets that demand complex reasoning over extended contexts, specifically: 
(1) \textbf{NarrativeQA} from the HELMET benchmark~\cite{yen2024helmet}, which consists of 1,272 questions with the longest document reaching 518k tokens. 
(2) \textbf{DetectiveQA}~\cite{xu2025detectiveqa} is a bilingual detective story dataset with an average length exceeding 100k tokens, requiring precise evidence localization across scattered plot points.
\paragraph{Long-term dialogue memory} 
We evaluate our model on \textbf{LoCoMo}~\cite{maharana2024evaluating}, a large-scale benchmark designed for long-term dialogue memory. The dataset comprises 50 multi-session dialogues across 10 distinct user groups, with each dialogue averaging approximately 9,000 tokens. Following prior work, we report performance across four fine-grained categories: \text{single-hop}, \text{multi-hop}, \text{temporal reasoning}, and \text{open-domain}. 

\subsection{Baselines}
We evaluate QRRanker against a broad spectrum of retrieval and memory frameworks.

For general-purpose reranking on Wikipedia QA and Long-context story  tasks, we compare QRRanker against two categories of models: 
(1) \textbf{Embedding Models}: Qwen3-Embedding (4B/8B)~\cite{zhang2025qwen3} and SFT-Embedding-8B, which is fine-tuned from Qwen3-Embedding-8B on our constructed data.
(2) \textbf{Reranking Methods}: HippoRAG~\cite{jimenez2024HippoRAG,gutierrez2025}, GroupRank~\cite{sun2025grouprank}, ReasonRank~\cite{liu2025reasonrank}, Qwen3-Reranker-4B (out-of-box)~\cite{zhang2025qwen3}, and a Qwen3-Reranker-4B variant trained on the same data as our QRRanker. We also include the \text{QRHead} without training as a baseline.

For the long-term dialogue task on LoCoMo, we compare QRRanker with a range of strong baselines, including: A-Mem~\cite{xu2025mem}, MemoryOS~\cite{li2025memos}, Zep~\cite{rasmussen2025zep}, Mem0~\cite{chhikara2025mem0}, Nemori~\cite{nan2025nemori}, and LightMem~\cite{fang2025lightmem}; TiMem~\cite{li2026timem}, Synapse~\cite{jiang2026synapse}, Membox~\cite{tao2026membox}, CompassMem~\cite{hu2026memory}, and ES-Mem~\cite{zou2026mem}; SimpleMem~\cite{liu2026simplemem}. Detailed baseline descriptions are provided in Appx.~\ref{appx:baselines}.

\textbf{Additional Comparisons.}\ 
Beyond the main benchmarks, we include two additional comparisons to examine QRRanker's scope. First, we evaluate QRRanker on BRIGHT~\citep{su2025bright}, a reasoning-intensive retrieval benchmark, and compare it with reasoning rerankers: ReasonRank and GroupRank. Second, we compare one-step QRRanker reranking with recent retrieval agents, DCI-Agent~\citep{li2026beyond} and Nemo Retriever Agent~\citep{nvidia2026nemoretriever}, to contrast lightweight reranking with iterative agentic retrieval. Details are provided in Appx~\ref{appx:bright} and~\ref{appx:agent}.
\subsection{Implementation Details}
\label{sec:implementation}
QRRanker is trained on \texttt{Qwen3}-\texttt{4B}-\texttt{Instruct}-\texttt{2507}. Unless otherwise stated, all results are based on
this backbone. We fine-tune the full backbone using a contrastive ranking loss computed from the attention scores of selected QR heads. To verify the generalizability of head selection and cross-architecture transferability, we additionally experiment with
\texttt{Llama-3.2-3B Instruct}. QR head selection details for both models are provided in
Appx.~\ref{appx:qrheads}.
During training, the $scale$ factor in the max-min norm is set to 8, the batch size is 1, the gradient accumulation step is 4, and the learning rate is $1\times 10^{-5}$. 
We adopt the DeepSpeed ZeRO Stage 2 strategy and use 8 NVIDIA H20 GPUs.

For downstream QA evaluation, we use task-specific prompting for generation; the full prompt templates for NarrativeQA, DetectiveQA, and LoCoMo are provided in Appx.~\ref{appx:prompt}. 
We employ \textbf{Qwen3-8B} as the generator for NarrativeQA and DetectiveQA, where books are chunked into non-overlapping passages of $\sim$200 tokens. 
For the LoCoMo benchmark, we utilize \textbf{GPT-4o-mini} and \textbf{GPT-5-mini} as the generators. We segment the dialogue history into small chunks, ensuring that utterance continuity is preserved, with an average chunk size of 258 tokens. When enabling the memory-aware setting, we prepend a summary prefix before the ranked chunk list. We cap the summary prefix at 512 tokens and select summaries based on their coverage of the retrieved/reranked chunks.

\section{Results}

\begin{table*}[t]
\centering
\scriptsize
\renewcommand{\arraystretch}{1}
 \setlength{\tabcolsep}{8pt}
 \scalebox{0.95}{
 \begin{tabularx}{\textwidth}{l c c c c c c c}
\toprule
\textbf{LLM} & \textbf{Method} & \textbf{Tokens} &
\textbf{Single-hop} & \textbf{Multi-hop} & \textbf{Temporal} &
\textbf{Open-domain} & \textbf{Overall F1} \\

\midrule
GPT-4o-mini & Qwen3-Emb-8B (out-of-box) & 846 & 47.95 & 35.24 & 41.36 & 24.79 & 42.81 \\
\rowcolor{gray!20}GPT-4o-mini & SFT-Emb-8B & 841 & 57.22 & 37.06 & 56.27 & 29.11 & 51.58 \\
GPT-4o-mini & A-Mem~\cite{xu2025mem}$^{\dagger}$ & 2,712 & 44.65 & 27.02 & 45.85 & 12.14 & 39.65 \\
GPT-4o-mini & MemoryOS~\cite{li2025memos}$^{\dagger}$ & 3,874 & 48.62 & 35.27 & 41.15 & 20.02 & 42.84 \\
GPT-4o-mini & Zep~\cite{rasmussen2025zep}$^{\dagger}$ & 3,911 & 49.56 & 35.74 & 42.00 & 19.37 & 43.56 \\
GPT-4o-mini & Mem0~\cite{chhikara2025mem0}$^{\dagger}$ & 1,764 & 47.65 & 38.72 & 48.93 & 28.64 & 45.09 \\
GPT-4o-mini & Nemori~\cite{nan2025nemori}$^{\dagger}$ & 4,767 & 46.33 & 32.36 & 55.99 & \text{29.19} & 44.72 \\
GPT-4o-mini & LightMem~\cite{fang2025lightmem}$^{\dagger}$ & 815 & 47.64 & 32.11 & 53.79 & 26.14 & 44.73 \\
GPT-4o-mini & TiMem~\cite{li2026timem} & 511 & -- & -- & -- & -- & 54.40 \\
GPT-4o-mini & Synapse~\cite{jiang2026synapse} & 814 & 48.90 & 35.70 & 50.10 & 25.90 & 40.50 \\
GPT-4o-mini & Membox~\cite{tao2026membox} & 2,166 & 60.09 & 39.88 & 58.03 & 27.96 & 53.10 \\
GPT-4o-mini & CompassMem~\cite{hu2026memory} & 20,000 & 57.36  &  38.84 & 57.96 & 26.61 & 52.18 \\
GPT-4o-mini & ES-Mem~\cite{zou2026mem}$^{\dagger}$ & 2,925 & 50.07 & 36.52 & 47.90 & 24.77 & 45.56 \\
GPT-4.1-mini & SimpleMem~\cite{liu2026simplemem} & 531 & 51.12 & \text{43.46} & 58.62 & 19.76 & 43.24 \\
\midrule
GPT-4o-mini & \textbf{QRRanker (Ours)} & 854 & \textbf{62.95} & \textbf{43.06} & \textbf{61.90} & \textbf{29.79} & \textbf{57.03} \\

GPT-5-mini & \textbf{QRRanker (Ours)} & 854 & \textbf{61.78} &\textbf{44.73} & \textbf{64.53} & \textbf{31.04} & \textbf{57.32} \\
\bottomrule
\end{tabularx}
 }
\caption{Comparison with SOTA Memory and Agent frameworks on the LoCoMo. Results marked with $\dagger$ are derived from ES-Mem~\cite{zou2026mem}. For QRRanker, we rerank the top-50 chunks retrieved by SFT-Emb-8B and utilize only the top-3 chunks as context for generation, without additional memory mechanisms. `--' indicates the metric is not reported in the corresponding paper.}
\label{tab:sota_memory_comparison_detailed}
\end{table*}

\subsection{Main Results}
We conduct experiments across three representative long-context settings: multi-hop question answering over Wikipedia, long-story question answering, and dialogue memory, covering five datasets in both English and Chinese. In our main evaluation, QRRanker is built upon the \texttt{Qwen3-4B-Instruct-2507} backbone, while we also provide results for a \texttt{Llama-3.2-Instruct} variant in Appx.~\ref{appx:llama_results} to demonstrate that QRRanker is robust across different backbones.
Tables~\ref{tab:longctx_recall_v2} and~\ref{tab:locomo_retrieval_results} summarize reranking performance in terms of Recall@$k$, while Tables~\ref{tab:sota_memory_comparison_detailed} and~\ref{tab:qa_performance_r10} report downstream generation results.
Overall, QRRanker consistently achieves the best overall results across settings, demonstrating improvements in both retrieval quality and downstream task performance.
\noindent\paragraph{Rerank Performance.}
We first analyze the effectiveness of QRRanker when applied to rerank the candidates retrieved by the first-stage retriever.

As shown in Table~\ref{tab:longctx_recall_v2}, QRRanker establishes a new state-of-the-art benchmark. It surpasses the Qwen-Reranker-4B by a substantial margin and improves the average recall significantly. On Wikipedia datasets such as Musique and HotpotQA, QRRanker outperforms complex graph-based methods like HippoRAG~\cite{gutierrez2025}. The performance gap is particularly evident in the story domain, where context tracking is critical. 
Under our long-context reranking evaluation, QRRanker achieves competitive or
stronger average recall than much larger 32B rerankers, such as
GroupRank~\cite{sun2025grouprank} and ReasonRank~\cite{liu2025reasonrank}. We further compare QRRanker with these listwise rerankers on BRIGHT in Appx.~\ref{appx:bright}, demonstrating its effectiveness in reasoning-intensive retrieval. Furthermore, QRRanker maintains this advantage on LoCoMo, despite using a substantially smaller backbone. These results suggest that QR-head training is an effective lightweight alternative for long-context reranking.


\noindent\paragraph{Long-context Story QA Performance.}

\begin{table}[t]
\centering
\scriptsize
\renewcommand{\arraystretch}{1}
\setlength{\tabcolsep}{7pt} 
\scalebox{0.95}{
\begin{tabularx}{\columnwidth}{l | cc | c} %
\toprule
\multirow{2}{*}{\textbf{Methods}} & \multicolumn{2}{c|}{\textbf{NarrativeQA}} & \textbf{DetectiveQA} \\
& F1 & EM & \textbf{ACC} \\
\midrule
\rowcolor[gray]{0.95} \multicolumn{4}{l}{\textit{Embedding Methods}} \\
Qwen3-Embedding-8B   & 26.30 & 11.01  & 57.35 \\
SFT-Embedding-8B     & 28.48 & 12.11  & 62.85 \\
\midrule
\rowcolor[gray]{0.95} \multicolumn{4}{l}{\textit{Reranking Methods}} \\
Qwen3-Reranker-4B (vanilla) & 29.10 & 12.58 & 60.93 \\
Qwen3-Reranker-4B (trained)  & 30.51 &13.52 & 64.52 \\
\midrule
\rowcolor[gray]{0.95} \multicolumn{4}{l}{\textit{QRRanker Series}} \\
QRHeads-4B           & 31.40 & 14.70 & 64.75 \\
\textbf{QRRanker} & \textbf{33.61} & \textbf{16.04} & \textbf{67.25} \\
\bottomrule
\end{tabularx}
}
\caption{QA performance on NarrativeQA and DetectiveQA. All methods utilize R@3 retrieved chunks as the context for generation (Qwen3-8B as Generator).}
\label{tab:qa_performance_r10}
\end{table}

High-quality retrieval should translate to improved generation accuracy. We evaluate this on narrative understanding datasets. As shown in Table~\ref{tab:qa_performance_r10}, QRRanker significantly improves downstream QA performance.  On NarrativeQA, it achieves 33.61 F1, outperforming the trained Qwen3-Reranker-4B (30.51). On DetectiveQA, accuracy increases from 62.85 (SFT-Embedding-8B) to 67.25 with QRRanker. 
These results suggest that QRRanker selects evidence that is not only semantically relevant, but also better aligned with the reasoning needed for answer generation.

\paragraph{Dialogue Memory Performance} As summarized in Table~\ref{tab:sota_memory_comparison_detailed}, QRRanker achieves the best Overall F1 on LoCoMo with a highly compact input budget. Using only 854 tokens on average (top-3 chunks) from the raw dialogue history, it attains 57.03 Overall F1 with GPT-4o-mini and 57.32 with GPT-5-mini. In contrast, many memory-augmented baselines require substantially larger token budgets to maintain explicit memory stores or graph structures. QRRanker instead reranks the top-50 chunks retrieved by the embedding retriever and feeds only a few top-ranked raw dialogue chunks to the generator.  This simple and lightweight design remains highly effective at capturing long-range dependencies, yielding the superior overall performance in our LoCoMo comparison.

\subsection{Results with Contextual Information}

\begin{table}[!]
    \centering    
    \small
    \renewcommand{\arraystretch}{1} 
    \setlength{\tabcolsep}{7pt}
    \scalebox{0.9}{
    \begin{tabular}{l ccc}
        \toprule
        \multirow{2}{*}{\textbf{Dataset}} & \multicolumn{3}{c}{\textbf{QRRanker}} \\
        \cmidrule(lr){2-4}
         & Chunk & +Sum & $\Delta$ \\
        \midrule
        LoCoMo      & 86.64 & 87.34 & \textcolor{poscolor}{\textbf{+0.70}} \\
        NarrativeQA & 28.09 & 29.11 & \textcolor{poscolor}{\textbf{+1.02}} \\
        DetectiveQA & 29.55 & 32.22 & \textcolor{poscolor}{\textbf{+2.67}} \\
        
        HotpotQA    & 95.05 & 94.75 & \textcolor{negcolor}{\textbf{-0.30}} \\
        Musique     & 70.19 & 70.16 & \textcolor{negcolor}{\textbf{-0.03}} \\
        
        \bottomrule
    \end{tabular}
    }
    
\caption{Recall@3 comparison of QRRanker with chunk-only inputs versus a summary prefix (+Sum) as contextual memory. }
\label{tab:summary_ablation}
\end{table}
As shown in Table~\ref{tab:summary_ablation}, equipping QRRanker with a summary prefix consistently improves ranking performance across long-dialogue and long-context story benchmarks. This suggests that the summary provides global contextual guidance, complementing the fine-grained evidence from retrieved chunks. 
Moreover, we test summary-based memory on Wikipedia-based multi-hop QA. We build a hierarchical clustering tree over retrieved passages and use parent summaries as the prefix. However, this strategy brings no gains and can even degrade performance, suggesting that abstracted global summaries are less helpful when evidence is highly localized in Wikipedia passages. 


\subsection{Heads from Different Layer- Levels}\label{head_exp}


QRRanker uses a fixed set of precomputed QR-heads, raising the question of which layer ranges provide suitable initialization for QR training. To study this, we use a semi-automatic variant that selects 16 heads from a continuous layer range $l_s$--$l_e$ for each sample. We evaluate QRRanker and its variants on NarrativeQA, with implementation details provided in Appx.~\ref{appx:variant}.

\begin{table}[ht]
\centering
\small
\renewcommand{\arraystretch}{1}
\scalebox{0.9}{
\begin{tabular}{l ccc}
\toprule
\textbf{Methods} & \textbf{R@3} & \textbf{R@5} & \textbf{R@10}\\
\midrule
QRRanker  & 28.87 & 39.16 & 54.44  \\
\midrule
10-17 & 24.51 & 34.52 & 49.91  \\
17-24 & 28.15 & 39.07 & 54.28  \\
28-35 & 28.48 & 38.88 & 54.65  \\
\bottomrule
\end{tabular}
}
\caption{Retrieval performance on NarrativeQA of QRRanker and its variants adapted on different levels of layers. $l_{s}-l_{e}$ denotes the layers with head selection.}
\label{tab:variant_performance}
\end{table}

As shown in Table~\ref{tab:variant_performance}, selecting heads from lower layers (10--17) leads to a clear performance drop, while selecting from middle layers (17--24) or higher layers (28--35) achieves performance close to QRRanker. This suggests that QR training is more effective when initialized from middle-to-high layers, where retrieval-relevant behavior is more likely to emerge. Since the 17--24 variant uses only middle-layer heads while preserving comparable performance, it also motivates a truncated middle-layer variant for more efficient inference.

Interestingly, the heads selected by the 17--24 variant have low overlap with the original QR-heads. This indicates that QR training can activate retrieval ability in diverse mid-layer heads, rather than relying exclusively on the originally discovered QR-heads. Nevertheless, QRRanker still performs slightly better, suggesting that precomputed QR-heads provide a stronger initialization prior. Further head-level comparisons are provided in Appx.~\ref{appx:head_validation}. We report the inference efficiency of the truncated middle-layer variant in \textsection~\ref{sec:efficiency}.

\subsection{Inference Efficiency}
\label{sec:efficiency}

We compare inference efficiency on 20 random samples.
Table~\ref{tab:latency_pctl_cost} shows that \texttt{QRRanker} reduces P50/P95 latency, TFLOPs, and peak memory compared with \texttt{Qwen3-Reranker-4B}. \texttt{QRRanker(middle)} further truncates the model after layer 24, achieving the lowest latency and resource cost. For \texttt{Qwen3-Reranker-4B}, we report both \emph{batch=50}, which processes all chunk--query pairs in one forward pass, and \emph{batch=1}, which processes them separately. These results show that QRRanker provides a more efficient reranking interface, especially with middle-layer truncation.

\subsection{Further Analysis}
\label{sec:further_analysis}

We provide further analyses to better understand QRRanker's behavior and generality.
QRRanker remains robust when the same top-50 candidate set is randomly shuffled on HotpotQA and LoCoMo, suggesting that its scores primarily reflect content-level relevance rather than positional cues (Appx.~\ref{appx:position_bias}). 
Moreover, we compare QRRanker with advanced retrieval agents, showing that it generalizes beyond standard reranking settings while retaining a simple one-step retrieval--reranking pipeline (Appx.~\ref{appx:agent}).

\begin{table}[t]
\centering
\small
 \renewcommand{\arraystretch}{1} 
\setlength{\tabcolsep}{3.5pt}
\scalebox{0.77}{
\begin{tabular}{lcccc}
\toprule
\multirow{2}{*}{\textbf{Method}} & \textbf{P50} & \textbf{P95} & \textbf{TFLOPs} & \textbf{Peak Mem} \\
& (ms) & (ms) & (/query) & (GB) \\
\midrule
Qwen3-Reranker (batch=50) & 1221.59 & 1256.29 &115.69 & 13.88 \\
Qwen3-Reranker (batch=1) &1895.26 &1929.09 &113.65 &\textbf{7.78} \\
\textbf{QRRanker}   &1095.42 &1133.38 &82.74 &11.18  \\
\textbf{QRRanker (middle)}   & \textbf{910.42} & \textbf{928.1} & \textbf{69.83} & \text{8.71} \\
\bottomrule
\end{tabular}
}
\caption{Inference efficiency comparison in latency, compute, and peak GPU memory. All models are evaluated under the same hardware and inference settings. \texttt{QRRanker(middle)} truncates the model after layer 24.
}
\label{tab:latency_pctl_cost}
\end{table}

\section{Conclusion}
In this paper, we present \textbf{QRRanker}, a lightweight and efficient listwise reranking framework built on Query-focused Retrieval (QR) heads in LLMs. By explicitly training selected QR heads for ranking, QRRanker produces real-valued relevance scores and performs reranking without generation at inference time.
Across five datasets spanning Wikipedia multi-hop QA, long-context story QA, and dialogue memory, QRRanker consistently improves reranking quality and downstream QA performance. QRRanker remains practical with a small backbone and offers clear inference efficiency benefits. Moreover, it supports simple extensions such as an optional summary prefix for global context and mid-layer head selection for further efficiency.


\section*{Limitations}
While QRRanker demonstrates strong performance across multiple domains and datasets, several limitations remain. First, although we validate QRRanker on two backbone architectures , its behavior on larger-scale models (e.g., 14B+) remains unexplored. Moreover, larger models with more attention heads per layer could offer a richer pool of retrieval-sensitive candidates, potentially improving both head selection quality and final reranking performance. We leave this exploration for future work. Second, part of our training supervision relies on silver evidence rather than fully human-annotated gold evidence, due to the lack of fine-grained evidence annotations in narrative-style QA benchmarks. This may introduce label noise, especially when partially relevant passages are not covered by the constructed evidence set. Nevertheless, the consistent improvements across datasets indicate that QRRanker remains reasonably robust under this realistic weak-supervision setting.

\bibliography{software}

\newpage
\appendix
\section{Construction of Listwise Training Instances}
\label{appx:build_train_instances}

As summarized in Alg.~\ref{alg:build_train_instances}, we construct each listwise training instance for NarrativeQA by first retrieving top-$K$ candidate chunks for a question, assigning binary labels based on silver evidence, and optionally prepending a de-duplicated summary prefix as global context.

\begin{algorithm}[t]
\caption{Construct listwise training instances on NarrativeQA with optional summary prefix}
\label{alg:build_train_instances}
\begin{algorithmic}[1]
\Require NarrativeQA training split $\mathcal{D}$;
retriever $\mathcal{R}$; top-$K$ ($K{=}50$);
memory flag $\mathsf{UseMem}$; summary map $\mathcal{M}$
\Ensure Training set $\mathcal{T}$

\State $\mathcal{T}\gets \emptyset$
\ForAll{question $Q$ in $\mathcal{D}$}
  \State $G \gets \textsc{SilverEvidence}(Q)$ \Comment{constructed following~\cite{li2025mindscape}}
  \State $C \gets \mathcal{R}(Q, K)$ \Comment{retrieve top-$K$ chunks}
  \ForAll{$c_i \in C$}
    \State $y_i \gets \mathbb{I}[c_i \in G]$
  \EndFor

  \If{$\mathsf{UseMem}$}
    \State $\mathcal{S} \gets \textsc{LookupSummaries}(C, \mathcal{M})$ \Comment{map chunks in $C$ to summaries}
    \State $M \gets \textsc{MergeDedup}(\mathcal{S})$ \Comment{merge \& de-duplicate summaries}
  \Else
    \State $M \gets \emptyset$
  \EndIf

  \State $\mathcal{T} \gets \mathcal{T} \cup \{(Q, M, C, \{y_i\}_{i=1}^{K})\}$
\EndFor
\State \Return $\mathcal{T}$
\end{algorithmic}
\end{algorithm}

\begin{table*}[]
    \centering
    \scalebox{0.78}{
    \begin{tabular}{l c ccccccc cc ccc} \toprule 
    & & \multicolumn{7}{c}{\textbf{Stackoverflow}} & \multicolumn{2}{c}{\textbf{Coding}} & \multicolumn{3}{c}{\textbf{Theorem-based}} \\ 
    \cmidrule(lr){3-9} \cmidrule(lr){10-11} \cmidrule(lr){12-14}
    \textbf{Model} & \textbf{Avg.} & \textbf{Bio.} & \textbf{Earth.} & \textbf{Econ.} & \textbf{Psy.} & \textbf{Rob.} & \textbf{Stack.} & \textbf{Sus.} & \textbf{Leet.} & \textbf{Pony} & \textbf{AoPS} & \textbf{TheoQ.} & \textbf{TheoT.} \\ \midrule 
    ReasonRank 7B$^{\dagger}$ & 32.5 & 51.6 & 43.4 & 32.4 & 44.0 & 31.0 & 25.6 & 39.8 & 15.4 & \underline{20.1} & 7.0 & 38.9 & 40.7 \\ 
    ReasonRank 32B$^{\dagger}$ & 35.6 & 53.9 & 47.6 & \underline{36.3} & \textbf{52.6} & \underline{36.5} & 34.2 & \textbf{44.5} & 15.2 & 14.8 & 5.5 & \underline{40.6} & \underline{45.3} \\ 
    GroupRank 7B$^{\dagger}$ & 34.3 & 52.7 & 51.0 & 33.8 & 44.5 & 32.1 & 33.9 & 38.1 & \underline{16.3} & 17.3 & \underline{8.7} & \textbf{40.7} & 42.4 \\ 
    GroupRank 32B$^{\dagger}$ & \textbf{38.0} & \underline{59.0} & \underline{57.5} & \textbf{39.2} & \underline{50.0} & \textbf{39.1} & \textbf{39.0} & \underline{42.7} & 14.3 & 14.9 & \textbf{12.6} & 39.0 & \textbf{48.8} \\ \midrule
    QRRanker (3B) & \underline{36.2} & \textbf{60.7} & \textbf{60.0} & 32.3 & 47.1 & 34.2 & \underline{37.8} & 35.6 & \textbf{30.2} & \underline{19.4} & 8.3 & 34.5 & 34.0 \\ \bottomrule 
    \end{tabular}
    }
    \caption{nDCG@10 results of QRRanker and reasoning rerankers on BRIGHT. $^{\dagger}$ denotes the results are cited from~\citet{sun2025grouprank}. QRRanker ranks in second place on average compared with reasoning rerankers.}
    \label{tab:bright_result}
\end{table*}

\section{Position Bias Analysis}
\label{appx:position_bias}
Since QRRanker concatenates all 50 candidates under causal attention, a natural question is whether it exploits positional cues rather than semantic relevance.
To test this, we keep the same top-$50$ candidate set for each query and evaluate three conditions:
(i)~the original retriever order preserved,
(ii)~a {randomly shuffled} order used directly, and
(iii)~QRRanker reranking on both the original and shuffled inputs.

\begin{table}[ht]
\centering
\small
\setlength{\tabcolsep}{4.5pt}
\renewcommand{\arraystretch}{1.0}

\begin{tabular}{l l c c c}
\toprule
\textbf{Dataset} & \textbf{Setting} & \textbf{R@3} & \textbf{R@5} & \textbf{R@10} \\
\midrule
\multirow{4}{*}{HotpotQA}
& Retriever (original)   & 82.85 & 89.05 & 95.15 \\
& Random (shuffled)      &  7.05 & 11.70 & 20.10 \\
& QRRanker (original)    & 95.05 & 96.90 & 97.70 \\
& QRRanker (shuffled)    & 91.65 & 95.20 & 97.00 \\
\midrule
\multirow{4}{*}{LoCoMo}
& Retriever (original)   & 76.01 & 83.10 & 90.15 \\
& Random (shuffled)      &  6.04 &  9.99 & 18.96 \\
& QRRanker (original)    & 87.34 & 91.32 & 95.01 \\
& QRRanker (shuffled)    & 85.35 & 90.60 & 93.78 \\
\bottomrule
\end{tabular}

\caption{Position bias analysis. ``Random (shuffled)'' denotes using a randomly permuted order directly as the ranking output without any reranking.}
\label{tab:position_bias}
\end{table}

As shown in Table~\ref{tab:position_bias}, the random-order baseline collapses to near-chance levels after shuffling, indicating that positional information is largely destroyed.
In contrast, QRRanker remains robust under shuffled input, with a maximum drop of only 3.40 R@3 on HotpotQA.
Notably, even after shuffling, QRRanker still substantially outperforms the original-order retriever, demonstrating that its ranking quality stems from content-level relevance matching rather than positional cues.
We further assess ranking stability by running five independent random permutations on LoCoMo.
The resulting Kendall's $\tau = 0.638 \pm 0.100$ and Spearman's $\rho = 0.808 \pm 0.096$ indicate high agreement across permutations, suggesting that QRRanker produces consistent rankings largely independent of order.

\section{Comparison on BRIGHT}
\label{appx:bright} 
Reasoning rerankers like GroupRank~\citep{sun2025grouprank} and ReasonRank~\citep{liu2025reasonrank} focus on reasoning-intensive retrieval. They are trained by a sophisticated pipeline with supervised finetuning and reinforcement learning to possess the think-then-rank ability. To verify QRRanker's capability of handling reasoning-intensive retrieval scenarios and compare it with these rerankers excelling in reasoning, we trained QRRanker using the open-source training set of ReasonRank~\citep{liu2025reasonrank}\footnote{\url{https://huggingface.co/datasets/liuwenhan/reasonrank_data_13k}}. The training process involves no reasoning CoT generation but only uses the positive/negative annotations for our contrastive learning.

We compare QRRanker with ReasonRank and GroupRank on the reasoning dataset BRIGHT~\citep{su2025bright}. Following GroupRank's setting, QRRanker also uses the original queries to rerank the top-100 passages retrieved by DIVER-Retriever-4B~\citep{DIVER} using GPT4-rewritten queries. The results are shown in Table~\ref{tab:bright_result}. QRRanker, a 3B model, is trained with the same data as ReasonRank and produces better results than ReasonRank's 32B version on average, indicating its inner reasoning ability may be activated by our QR training without complicated thinking generation. As for GroupRank, the model is trained with closed-source and more diverse data constructed by the authors. Additionally, GroupRank's generation is performed several times to get the self-consistency results, where self-consistency is proven to improve CoT reasoning by~\citet{wang2023selfconsistency}. QRRanker can defeat GroupRank 7B and show comparable performance against GroupRank 32B, which demonstrates QRRanker's potential in reasoning-intensive retrieval with easy training adaptation.


\section{One-step QRRanker Reranking v.s. Advanced Retrieval Agents}\label{appx:agent}

\begin{table*}[t]
    \centering
    \scalebox{0.72}{
    \begin{tabular}{l cc cc c ccccc}
    \toprule
    & \multicolumn{5}{c}{Recall@3/5/10} & \multicolumn{5}{c}{nDCG@10} \\
    \cmidrule(lr){2-6} \cmidrule(lr){7-11}
    & \multicolumn{2}{c}{\textbf{StoryQA}} & \multicolumn{2}{c}{\textbf{WikipediaQA}} & \textbf{Conv.} & \multicolumn{5}{c}{\textbf{BRIGHT}} \\
    \cmidrule(lr){2-3} \cmidrule(lr){4-5} \cmidrule(lr){6-6} \cmidrule(lr){7-11}
    \textbf{Model} & NarQA & DetQA & MuSiQue & HotpotQA & LoCoMo & Avg. & Bio. & Earth. & Econ. & Rob. \\
    \midrule
    DCI-Agent & 12.7/15.6/19.3            & 22.0/28.0/30.0      & 63.6/68.3/75.3            & 97.0/98.5/98.5             & 86.3/88.6/91.4            & 46.4 & 60.0$^{*}$ & 50.8$^{*}$ & 32.3$^{*}$ & \textbf{42.4}$^{*}$ \\
    Nemo Agent     & 27.3/\textbf{36.5}/\textbf{47.4} & 23.0/35.0/42.0    & \textbf{76.9}/\textbf{86.7}/\textbf{92.1} & \textbf{98.0}/\textbf{99.5}/\textbf{99.5} & \textbf{87.5}/\textbf{90.5}/\textbf{96.5} & 45.7 & 55.7 & 51.5 & \textbf{35.6} & 39.8 \\
    QRRanker   & \textbf{28.0}/36.4/47.2 & 2\textbf{5.0}/\textbf{39.0}/\textbf{49.0}    & 67.2/75.1/78.8 & 93.5/95.5/97.0 & 86.8/90.2/96.3 & \textbf{46.8} & \textbf{60.7} & \textbf{60.0} & 32.3 & 34.2 \\
    \bottomrule
    \end{tabular}
    }
    \caption{Performance comparison across dataset groups. StoryQA includes NarrativeQA and DetectiveQA (averaged over EN and ZH); WikipediaQA includes MuSiQue and HotpotQA; Conv. denotes LoCoMo; BRIGHT includes Biology, Earth Science, Economics, and Robotics subsets according to~\citet{li2026beyond}. $^{*}$ denotes the results are cited from~\citet{li2026beyond}.}
    \label{tab:agent_main_results}
\end{table*}

\begin{table}[t]
    \centering
    \scalebox{0.69}{
    \begin{tabular}{l c l c c c c}
    \toprule
    \textbf{Dataset} & \textbf{\#C.(T)} & \textbf{Method} & \textbf{\#C.(R)} & \textbf{\#Ret.} & \textbf{\#Ti.(s)} & \textbf{\#Tok.} \\
    \midrule
    \multirow{3}{*}{Story}
        & \multirow{3}{*}{831}
        & Nemo & 242 & 12 & 97  & 280k \\
        & & DCI  & 111  & 16 & 41.7  & 57k   \\
        & & Ours & 50  & 1 & 1.4  & 10k  \\
    \midrule
    \multirow{3}{*}{LoCoMo}
        & \multirow{3}{*}{71}
        & Nemo & 71  & 13 & 89  & 205k \\
        & & DCI  & 37 & 16 & 32.7 & 27k   \\
        & & Ours & 50  & 1 & 1.8  & 13k   \\
    \midrule
    \multirow{3}{*}{Wiki}
        & \multirow{3}{*}{11k}
        & Nemo & 401 & 19 & 115 & 602k \\
        & & DCI  & 7  & 15 & 34.6  & 55k   \\
        & & Ours & 50  & 1 & 1.1  & 8k  \\
    \midrule
    \multirow{3}{*}{BRIGHT}
        & \multirow{3}{*}{73k}
        & Nemo & 239 & 12 & 61  & 168k \\
        & & DCI  & --  & -- & --  & --   \\
        & & Ours & 100  & 1 & 1.5  & 27k   \\
    \bottomrule
    \end{tabular}
    }
    \caption{Average resource consumption per sample across dataset groups. All values prefixed with \# denote averages. \textbf{C.(T)}: total chunks in the corpus; \textbf{C.(R)}: number of chunks retrieved; \textbf{Ret.}: number of times the retrieve function (or tool) is called; \textbf{Ti.(s)}: latency for finishing a sample in seconds; \textbf{Tok.}: total tokens consumed per sample. Because we cite the BRIGHT results of DCI-Agent from~\citet{li2026beyond}, we do not display its consumption on BRIGHT. In the table, we can see that Nemo Agent significantly invokes more retrieval calls in Wikipedia than in others. Due to \texttt{grep} tools reading localized and partial contents in a chunk, DCI-Agent refers to dozens of chunks but consumes significantly fewer tokens than Nemo.}
    \label{tab:resource_consumption}
\end{table}

We also compare QRRanker with the recent advanced retrieval agents: DCI-Agent~\citep{li2026beyond} and Nemo Retrieval Agent~\citep{nvidia2026nemoretriever}. DCI-Agent discards any similarity-based retrieval tools, and iteratively utilizes only \texttt{grep}/shell commands/read tools to access texts in chunks. Constructed for QA, DCI-Agent also achieves excellent performance in retrieval tasks (\textit{e.g.,} BRIGHT and BEIR subsets)~\citep{li2026beyond}. To this end, we extend DCI-Agent to StoryQA and WikipediaQA using its settings on BRIGHT. As for Nemo Agent, it is a typical embedding-retrieval-based ReAct~\citep{yao2023react} agent that iteratively recalls top-N chunks using embedding tools, and finally picks and reranks the required number of chunks among all retrieved chunks. Unlike these agents, we adapt QRRanker to a standard one-step pipeline: embedding retrieval $\rightarrow$ rerank, which can be a step of tool calling.

Following the settings of DCI-Agent, the LLM used by agents is GPT-5.4-nano with highly thinking; the context management of DCI-Agent, dealing with the over-length situation, shortens the large tool results in each turn, keeps recent turns, and replaces older tool results with placeholders (called Level3 by \citet{li2026beyond}). Nemo Agent uses a different context management: no truncation; when retrieved chunks explode the context, it ends the main agent and uses a selection agent to do the final reranking. To ensure fairness, we use Qwen3-Embedding-8B as the embedding model for Nemo Agent and our method on Story/Wikipedia/Conversation QA, Diver-Retriever-4B on BRIGHT. As agents are notoriously slow and expense-unfriendly, we randomly select 100 samples for Story/Wikipedia/Conversation QA, and BRIGHT aligns with DCI-Agent settings.

The results are shown in Table~\ref{tab:agent_main_results}, where Nemo Agent exhibits strong performance due to its harness engineering to fully utilize LLM and frequent retrieval. It is worth noting that QRRanker is competitive against retrieval agents as it achieves the best results on BRIGHT and StoryQA on average without any LLM interactive synergy. As for WikipediaQA, which is thoroughly studied with massive high-quality training resources, retrieval agents like Nemo Agent can significantly reason and trigger more retrieval-tool calls, which can be seen in Table~\ref{tab:resource_consumption}. Therefore, they almost kill this type of task, especially indicated by the results on HotpotQA. As for Nemo Agent defeating QRRanker on LoCoMo, Nemo Agent recalls and reranks all chunks in LoCoMo, showing in Table~\ref{tab:resource_consumption}. Referring to tasks like StoryQA and BRIGHT, which require integrative reasoning over dispersed evidence rather than compositional entity-hop resolution, Nemo Agent basically falls back to the same level as QRRanker. It indicates that current retrieval agents may need to enhance their capacity for holistic evidence synthesis. In the future, we will try to synergize the agent with our QRRanker to reach such a goal.


DCI-Agent shows a notable performance disparity across task types. On WikipediaQA, it achieves strong retrieval quality while introducing remarkably few chunks into the LLM context, largely because explicit entity names in the questions can often be matched precisely with \texttt{grep}. However, approximately 20\% of WikipediaQA queries issue \texttt{grep} commands containing terms from the \textit{gold answer} that are absent from the question, suggesting that the agent may leverage the LLM's parametric knowledge of Wikipedia content to guide retrieval. In contrast, on StoryQA, where evidence is distributed across long narratives and deeper comprehension is required, DCI-Agent's keyword-matching strategy degrades substantially, and it underperforms QRRanker and Nemo Agent with conventional retrieval tools. This phenomenon aligns with the findings of the DCI-Agent paper~\citep{li2026beyond}: \texttt{grep} verifies exact and localized spans in chunks instead of the whole chunk, leading to a lower recall but deeper exploitation of fine-grained discoveries~\citep{li2026beyond}. This utilization makes DCI-Agent a great question answerer. In future work, we will explore combining QRRanker with local search techniques to support deeper and fine-grained evidence mining beyond expanding retrieval coverage. 

Overall, QRRanker is competitive against retrieval agents, which does not imply QRRanker should replace agents. Rather, we believe that QRRanker can be a good enhancer, acting as a tool equipped for these agents, to reach a new level of retrieval coverage and deeper evidence digging. We leave this to future work, as we have mentioned above.

\section{LoCoMo Baselines}
\label{appx:baselines}
We compare QRRanker with a set of memory-augmented baselines on LoCoMo. Below, we provide brief descriptions of each method.

\begin{itemize}

    \item \textbf{TiMem}~\cite{li2026timem}: Organizes memories with a temporal hierarchical structure to retrieve long-horizon information efficiently.
\end{itemize}
\begin{itemize}
    \item \textbf{SimpleMem}~\cite{liu2026simplemem}: Compresses dialogue history into compact semantic memory to reduce redundancy and context length.
\end{itemize}

\begin{itemize}
    \item \textbf{SYNAPSE}~\cite{jiang2026synapse}: Models memory as a dynamic graph and retrieves relevant items via spreading activation.
\end{itemize}

\begin{itemize}
    \item \textbf{CompassMem}~\cite{hu2026memory}: Segments interactions into events and constructs an event-level structure to guide retrieval and reasoning.
    \item \textbf{ES-Mem}~\cite{zou2026mem}: Uses event segmentation to build coherent long-term memories for dialogue agents.
    \item \textbf{Membox}~\cite{tao2026membox}: Packs dialogue into topic-consistent memory units to preserve topic continuity over long contexts.
    \item \textbf{Mem0}~\cite{chhikara2025mem0}: A “memory-centric” architecture that dynamically extracts, integrates, and retrieves important information from conversations to build and maintain a scalable long-term memory.
    \item \textbf{Nemori}~\cite{nan2025nemori}: It employs a Two-Step Alignment Principle to structure dialogue streams into semantically coherent event segments and utilizes a Predict-Calibrate Principle to actively learn from prediction discrepancies, enabling the adaptive evolution of knowledge.
     \item \textbf{MemoryOS}~\cite{li2025memos}: An OS-inspired AI memory system featuring a hierarchical architecture with storage, updating, retrieval, and generation modules. It optimizes dynamic updates through FIFO dialogue chains and heat-based segmented paging.
    \item \textbf{Zep}~\cite{rasmussen2025zep}: Leveraging a dynamic and temporal-aware Knowledge Graph engine, it integrates unstructured dialogue data with structured business data while preserving their historical relationships.
    \item \textbf{LightMem}~\cite{fang2025lightmem}: A cognitively inspired architecture featuring sensory and short-term modules for lightweight compression and integration. Uniquely, it updates long-term memory during ``sleep time'' to decouple consolidation from online reasoning, balancing performance and efficiency.
\end{itemize}

\section{QR Head Selection}
\label{appx:qrheads}   
\paragraph{Qwen3-4B-Instruct-2507.}              
This model contains 36 layers of 32-head self-attention. We use 1000 random samples from the NarrativeQA training split as the seed set to compute QR scores and select the top-16 heads. The selected QR heads (denoted as $l$\texttt{-}$h$, where $0\leq l<36$ is the layer and $0\leq h<32$ is the head index) are: \texttt{20-15}, \texttt{21-11}, \texttt{17-27}, \texttt{23-10}, \texttt{22-4}, \texttt{21-10}, \texttt{21-8}, \texttt{21-18}, \texttt{18-15}, \texttt{18-19}, \texttt{17-25}, \texttt{17-17}, \texttt{24-13}, \texttt{17-4}, \texttt{19-12}, \texttt{21-31}. 

\paragraph{Llama-3.2-3B-Instruct.}
To verify that QR head selection generalizes across model architectures and seed datasets, we apply the same procedure to \texttt{Llama-3.2-3B-Instruct}  using 1000 random samples from the MuSiQue training split as the seed set. The selected QR heads are: \texttt{12-1}, \texttt{13-9}, \texttt{13-23}, \texttt{12-16}, \texttt{8-8}, \texttt{13-16}, \texttt{12-6}, \texttt{13-22}, \texttt{12-2}, \texttt{15-22}, \texttt{12-3}, \texttt{15-18}, \texttt{12-10}, \texttt{14-2}, \texttt{21-20}, \texttt{9-23}.

\section{Cross-Architecture Results with Llama-3.2-3B-Instruct}
\label{appx:llama_results} 
To verify cross-architecture generalization, we instantiate QRRanker on Llama-3.2-3B-Instruct. Table~\ref{tab:llama32_results} shows that this variant consistently outperforms both the base retriever and pretrained QRHeads across out-of-domain dataset: HotpotQA, LoCoMo, and DetectiveQA, demonstrating effective transfer of the QR training objective to different model families.
The results indicate that QRRanker transfers well across model architectures. Compared with the first-stage retriever, the Llama-3.2-3B-Instruct variant brings consistent gains on all three out-of-domain datasets, especially on LoCoMo and DetectiveQA where the initial retrieval quality is relatively weaker and reranking provides larger improvements. QRRanker also consistently surpasses QRHeads at all recall cutoffs. These results suggest that the QR training objective is not tied to a specific backbone architecture, and can effectively improve both  multi-hop retrieval and long-context narrative retrieval under out-of-domain transfer.

\begin{table}[t]
  \centering
  \small
  \setlength{\tabcolsep}{4pt}
  \renewcommand{\arraystretch}{1.1}
  \begin{tabular}{l ccc}
  \toprule
  \textbf{Dataset / Method} & \textbf{R@3} & \textbf{R@5} & \textbf{R@10} \\
  \midrule
  \multicolumn{4}{l}{\textit{HotpotQA (out-of-domain)}} \\
  Retriever & 82.85 & 89.05 & 95.15 \\
  QRHeads  & 89.85 & 93.75 & 96.40 \\
  \textbf{QRRanker } & \textbf{93.05} & \textbf{96.20} & \textbf{97.60} \\
  \midrule
  \multicolumn{4}{l}{\textit{LoCoMo (out-of-domain)}} \\
  Retriever & 58.61 & 67.57 & 79.15 \\
  QRHeads & 82.75 & 87.84 & 92.52 \\
  \textbf{QRRanker} & \textbf{84.27} & \textbf{88.98} & \textbf{93.65} \\
  \midrule
  \multicolumn{4}{l}{\textit{DetectiveQA (out-of-domain, avg EN/ZH)}} \\
  Retriever & 12.84 & 20.03 & 31.17 \\
  QRHeads  & 20.03 & 27.55 & 39.07 \\
  \textbf{QRRanker} & \textbf{22.04} & \textbf{28.88} & \textbf{40.65} \\
  \bottomrule
  \end{tabular}
  \caption{Out-of-domain results with Llama-3.2-3B-Instruct. The first-stage retriever is Qwen3-Embedding-8B. DetectiveQA scores are averaged over English and Chinese sets.}
  \label{tab:llama32_results}
  \end{table}

\section{Validation of QR Head Selection}
\label{appx:head_validation}

In \textsection~\ref{head_exp}, we show that heads from middle-to-top layers can reach comparable performance after QR training using the semi-auto selection variant. This raises a natural question: does the QR head discovery itself matter?
Table~\ref{tab:head_selection_validation} addresses this by comparing different head selection strategies on HotpotQA using the pretrained Qwen3-4B model, without any QR training.

\begin{table}[t]
\centering
\small
\setlength{\tabcolsep}{4pt}
\renewcommand{\arraystretch}{1.1}
\begin{tabular}{l ccc}
\toprule
\textbf{Method} & \textbf{R@1} & \textbf{R@3} & \textbf{R@5} \\
\midrule
Retriever (Qwen3-Emb-8B) & 45.45 & 82.85 & 89.05 \\
\midrule
\multicolumn{4}{l}{\textit{Pretrained (no QR training)}} \\
Random mid-layer (seed 42) & 43.45 & 84.85 & 92.55 \\
Random mid-layer (seed 123) & 41.70 & 84.10 & 93.35 \\
Random mid-layer (seed 999) & 40.75 & 81.15 & 91.60 \\
Discovered QR Heads & \textbf{45.50} & \textbf{90.20} & \textbf{94.80} \\
\midrule
\multicolumn{4}{l}{\textit{After QR training}} \\
QRRanker & \textbf{48.35} & \textbf{95.05} & \textbf{96.90} \\
\bottomrule
\end{tabular}
\caption{Head selection validation on HotpotQA (Qwen3-4B). Random mid-layer heads are sampled from layers 17--24.}
\label{tab:head_selection_validation}
\end{table}

Without any QR training, the discovered QR heads already outperform random mid-layer heads by +6.83 R@3 on average, indicating that the QR score selects heads with inherently stronger retrieval behavior. After QR training, performance further increases to 95.05 R@3, highlighting the complementary benefits of informed head selection and subsequent training.

\section{Variant with Semi-Auto Head Selection}
\label{appx:variant}

QRRanker statically trains and utilizes a group of precomputed QR heads. If we use a set of seed samples from another task to recompute QR scores, the QR heads may be different from the current ones. Our initial motivation for using the precomputed QR heads is that they provide a proper initialization. Along with training, heads will be forced to learn such a retrieval ability. We are curious about which part of heads are better suited to be a good starter, as QR heads do. Therefore, we propose a variant of QRRanker with semi-automatic head selection, which is limited to selecting heads from a local range of layers, but is free to choose heads from every layer for every sample. 

We set layers for head selection ranged from $l_{s}$ to $l_{e}$, where $0<l_{s}<l_{e}\leq36$. We restrict that the number of selected heads must equal 16 (the number of QR heads), and therefore, for simplified control, the model should select $n=16/(l_{e}-l_{s})$ heads per layer. To achieve selection, we follow the router technique of Mixture-of-Expert~\citep{moe} and add a gate to these layers. Instead of choosing MLPs for every token, our gate chooses $n$ heads for a sample. For selecting heads, we concatenate a repeat question $Q^{'}=[think]Q[/think]$ after the original question $Q$, where $Q^{'}$ is used for head selection and $Q$ is still for score computing. A gate of layer $l_{i}$ is a linear map from the dimension $32*d_{h}$ to $32$, with the trainable parameter $W_{l_{i}}\in \mathbb{R}^{d\times 32}$. The head score is computed by:
\begin{eqnarray}
    S_{l_{i}}=q_{l_{i}}\cdot W_{l_{i}},\\
    S_{l_{i}}=\text{mean}(\text{softmax}(S_{l_{i}}),\text{d}=0),
\end{eqnarray}
where $q_{l_{i}}\in \mathbb{R}^{|Q^{'}|\times d}$ is the hidden states of tokens in $Q^{'}$ at layer $l_{i}$, $d$ is the dimension of the hidden state, \text{cat}($\cdot$) is concatenating all query states along the head, mean($\cdot$, d=0) is averaging the score along the number of tokens in $Q^{'}$, and $S_{l_{i}}\in \mathbb{R}^{32}$ is the head score. We then choose the top-$n$ highest head scores $S_{l_{i}}^{Q}=[s_{h0}^{l_{i}},...,s_{hn}^{l_{i}}]$ and the corresponding heads. Following MoE, $S_{l_{i}}^{Q}$ is normalized to 1. After picking up heads for all layers with gates, these heads participate in computing retrieval scores, and the retrieval score will be multiplied by its head score $S_{l_{i}}^{Q}[x],0<x<n$ for the purpose of backward gradients. These gates will learn to select heads for samples during the QR training. 

In \textsection~\ref{head_exp}, we train QRRanker and the variant with training data only from NarrativeQA and evaluate them using the evaluation set of NarrativeQA. The training hyperparameters are set to the same as those in \textsection~\ref{sec:implementation}. We explore layers that can be used to select and train QR-like heads.

\section{Prompt Templates}
\label{appx:prompt}

\subsection{Block-based Summary Generation Prompt}
\label{appx:prompt_block_sum}

\begin{promptbox}
You are an expert fiction editor and continuity supervisor.

You are provided with a raw text segment from a book (Part \Var{sub\_idx} / \Var{total\_subs}).
This segment consists of approximately 20 consecutive chunks combined.

\texttt{<Raw\_Text>}\\
\Var{raw\_text}\\
\texttt{</Raw\_Text>}

Please generate a \textbf{Detailed Narrative Summary} following these strict guidelines:
\begin{enumerate}[label=\arabic*., nosep]
  \item \textbf{Narrative Reconstruction}: Do not list events. Rewrite the content as a coherent story in the third person, past tense. It should read like a condensed version of the original text.
  \item \textbf{Detail Preservation}:
  \begin{itemize}
    \item Preserve specific \textbf{Character Names} and their relationships.
    \item Keep key \textbf{Dialogues} that drive the plot.
    \item Note specific \textbf{Locations} or setting changes.
  \end{itemize}
  \item \textbf{Noise Filtering}:
  \begin{itemize}
    \item IGNORE any copyright notices, project gutenberg headers, page numbers, or table of contents.
    \item If the text starts or ends in the middle of a sentence, ignore the broken fragments and focus on the complete thoughts.
  \end{itemize}
  \item \textbf{Style}:
  \begin{itemize}
    \item NO meta-commentary (\emph{e.g.}, do NOT say ``The text describes...'', ``In this chunk...'').
    \item Directly tell the story.
  \end{itemize}
  \item \textbf{Length}: 50-100 words.
\end{enumerate}

Output the summary directly.
\end{promptbox}

\subsection{Event-centric Summary Generation}
\label{appx:prompt_event_sum}
\begin{promptbox}
\PromptLabel{Instruct:}
You are a specialized system for extracting structured event representations from conversational data.

\textbf{1. EVENT CLASSIFICATION}

\textbf{ANCHOR Events.}
Anchor events are MAJOR LIFE MILESTONES that will be remembered for years. Only classify as an anchor if the event meets ALL of these criteria:
\begin{itemize}
  \item Represents a first-time or rare life occurrence
  \item Has a lasting impact on the person's identity, relationships, or life trajectory
  \item Would be mentioned when telling someone about ``important moments in my life''
\end{itemize}
\textit{Examples of TRUE ANCHOR events:} First time attending LGBTQ support group, Starting adoption process, Career change, Moving to a new country, etc.

\textbf{EPHEMERAL Events.}
Most events are ephemeral. These include:
\begin{itemize}
  \item Plans and intentions (``I plan to...'', ``I want to...'')
  \item Routine activities (exercise, hobbies, daily tasks)
  \item Casual conversations and updates
  \item Past events being recalled (unless first mention of major milestone)
\end{itemize}

\textbf{4. RAW DIALOGUE REFERENCE}
\begin{itemize}
  \item related\_line\_indices: list the 2-4 most relevant line numbers (1-indexed from the dialogue)
  \item These lines will be saved as the event's source evidence
\end{itemize}

\PromptLabel{Input Dialog:} \Var{dialog}

\PromptLabel{Output (JSON format, extract 1-3 events):}
\begin{PromptJSON}
{
  "events": [
    {
      "summary": "concise description",
      "related_line_indices": [1, 2, 3]
    }
  ]
}
\end{PromptJSON}
\end{promptbox}

\subsection{QRRanker Instruction Template}
\label{appx:prompt_inst}
\begin{promptbox}
\PromptLabel{Instruct:}

\textbf{[Optional Memory Prefix $M$]}
Here are some session summaries that may help answer the query:
\Var{mapped summaries from top-50 chunks}

\textbf{[Candidate Chunks $C$]}
Here are some retrieved chunks:
\begin{enumerate}[label={[}\arabic*{]}, nosep]
  \item \Var{chunk $c_1$}
  \item \Var{chunk $c_2$}
  \item \Var{chunk $c_3$}
  \item \ldots
  \item \Var{chunk $c_{50}$}
\end{enumerate}

\PromptLabel{Query $Q$:} \Var{question }
\end{promptbox}

\subsection{LoCoMo QA Prompt}

\begin{promptbox}
You are an intelligent memory assistant tasked with retrieving accurate information from conversation memories.

\PromptLabel{Context:}
You have access to memories from two speakers in a conversation. These memories contain timestamped information that may be relevant to answering the question.

\PromptLabel{Instructions:}
\begin{enumerate}[label=\arabic*., nosep]
  \item Carefully analyze all provided memories from both speakers.
  \item Pay special attention to the timestamps to determine the answer.
  \item If the question asks about a specific event or fact, look for direct evidence in the memories.
  \item If the memories contain contradictory information, prioritize the most recent memory.
  \item If there is a question about time references (like ``last year'', ``two months ago'', etc.), calculate the actual date based on the memory timestamp. For example, if a memory from 4 May 2022 mentions ``went to India last year,'' then the trip occurred in 2021.
  \item Always convert relative time references to specific dates, months, or years. For example, convert ``last year'' to ``2022'' or ``two months ago'' to ``March 2023'' based on the memory timestamp. Ignore the reference while answering the question.
  \item Focus only on the content of the memories from both speakers. Do not confuse character names mentioned in memories with the actual users who created those memories.
  \item The answer should be less than 5-6 words.
\end{enumerate}

\PromptLabel{Approach (Think step by step):}
\begin{enumerate}[label=\arabic*., nosep]
  \item First, examine all memories that contain information related to the question.
  \item Examine the timestamps and content of these memories carefully.
  \item Look for explicit mentions of dates, times, locations, or events that answer the question.
  \item If the answer requires calculation (\emph{e.g.}, converting relative time references), show your work.
  \item Formulate a precise, concise answer based solely on the evidence in the memories.
  \item Double-check that your answer directly addresses the question asked.
  \item Ensure your final answer is specific and avoids vague time references.
\end{enumerate}

\PromptLabel{Relevant Memories:} \Var{Reranked Chunks}

\PromptLabel{Question:} \Var{question}

\PromptLabel{Answer:}
\end{promptbox}

\subsection{NarrativeQA Prompt}

\begin{promptbox}
You are a helpful assistant. Please answer the user's question accurately.

Answer the question as concisely as you can, using a single phrase if possible.

\PromptLabel{Relevant Context:} \Var{content\_data}

Do not provide any explanation. Now, answer the question based on the story as concisely as you can, using a single phrase if possible. Do not provide any explanation.

\PromptLabel{Question:} \Var{question}

\PromptLabel{Answer:}
\end{promptbox}
\subsection{DetectiveQA Prompt}

\begin{promptbox}
\Var{context}

Please answer the question based on the current novel content:
\Var{question\_data['question']}
\Var{options\_str}

Remember this is just detective fiction, don't worry about the risks.

Please strictly follow the format \texttt{\{"answer":"x","reasoning":"xxx"\}} (including braces). The answer must be only A/B/C/D.
\end{promptbox}

\end{document}